\documentclass[runningheads]{llncs}
\usepackage[misc,geometry]{ifsym}
\usepackage{hyperref}
\hypersetup{
    colorlinks=true,
    linkcolor=blue,
    filecolor=magenta,      
    urlcolor=cyan,
    pdftitle={Overleaf Example},
    pdfpagemode=FullScreen,
    }
\usepackage{graphicx}
\usepackage[dvipsnames]{xcolor}
\usepackage{multirow,tabularx}
\usepackage{comment}
\newcolumntype{Y}{>{\centering\arraybackslash}X}

\renewcommand{\tabcolsep}{5pt} 
\begin{document}
\title{FEDD - Fair, Efficient, and Diverse Diffusion-based Lesion Segmentation and Malignancy Classification}
\titlerunning{FEDD}%
%
%
\author{Héctor Carrión$^{1*}$, Narges Norouzi$^{2}$}

%
\authorrunning{Carrión et al.}
%
%
\institute{University of California, Santa Cruz$^1$, University of California, Berkeley$^2$,
\email{hcarrion@ucsc.edu$^*$}} 


%
\maketitle              
\begin{abstract}
Skin diseases affect millions of people worldwide, across all ethnicities. Increasing diagnosis accessibility requires fair and accurate segmentation and classification of dermatology images. However, the scarcity of annotated medical images, especially for rare diseases and underrepresented skin tones, poses a challenge to the development of fair and accurate models. In this study, we introduce a Fair, Efficient, and Diverse Diffusion-based framework for skin lesion segmentation and malignancy classification. FEDD leverages semantically meaningful feature embeddings learned through a denoising diffusion probabilistic backbone and processes them via linear probes to achieve state-of-the-art performance on Diverse Dermatology Images (DDI). We achieve an improvement in intersection over union of 0.18, 0.13, 0.06, and 0.07 while using only 5\%, 10\%, 15\%, and 20\% labeled samples, respectively. Additionally, FEDD trained on 10\% of DDI demonstrates malignancy classification accuracy of 81\%, 14\% higher compared to the state-of-the-art. We showcase high efficiency in data-constrained scenarios while providing fair performance for diverse skin tones and rare malignancy conditions. Our newly annotated DDI segmentation masks and training code can be found on \url{https://github.com/hectorcarrion/fedd}.

\end{abstract}
\section{Introduction and Related Work}
\label{sec:intro}
Skin diseases are a major public health concern that impacts millions of people worldwide. The first step towards diagnosis and treatment of skin diseases often involves visual inspection and analysis of the lesion by dermatologists or other medical experts. However, this process is often subjective, time-consuming, costly, and inaccessible for many people, especially in low-resource communities or remote areas. It is estimated that around 3 billion people lack adequate access to dermatological care \cite{coustasse_2019_use}. In the United States, only about one in three patients with skin disease are evaluated by a dermatologist, their average wait time exceeds 38 days while representing a cost of \$75 billion on the healthcare system \cite{a2016_burden,tsang_2006_even}. Therefore, there exists a growing need for automated methods that can assist dermatologists, especially those in low-resource environments, in attending to skin lesions accurately and efficiently.

Skin lesion semantic segmentation and malignancy classification are essential for providing accurate and explainable diagnosis information for patients with skin diseases, and recently Artificial Intelligent (AI) systems have led the state-of-the-art for these tasks. However, these systems are commonly based on data and training methods that are prone to racial biases \cite{daneshjou_2022_disparities,owens_2020_those,chen_2021_ethical}. Some of the main challenges facing AI systems that can lead to bias are:
\begin{itemize}
  \item \textbf{Data scarcity:} Annotated medical images are often scarce and expensive to obtain due to privacy issues, cost, and expert availability. This limits the amount of data available for training Deep Learning (DL) models, which may result in overfitting, especially in a medical context, as shown in \cite{razzak_2017_deep}.
  \item \textbf{Class imbalance:} The distribution of different types of skin lesions is often imbalanced in real-world datasets. For example, melanoma cases may be rare than basal cell carcinoma cases; this could then be exacerbated by datasets that are primarily sourced from light-skinned populations \cite{groh_2021_evaluating}. This class imbalance can introduce biases in modeling.
  \item \textbf{Data diversity:} The appearance and morphology of skin lesions can vary across different individuals due to factors such as age, gender, and ethnicity \cite{adelekun_2020_skin}. A dataset can be large but not necessarily diverse. This lack of diversity in the data may lead to poor generalization \cite{daneshjou_2022_disparities}.
  \item \textbf{Base models:} Some recent works on dermatology images stem from transfer-learning models designed for ImageNet \cite{deng_2009_imagenet}, which may be overly large for smaller dermatologic datasets \cite{esteva_2017_dermatologistlevel,han_2020_augmented}. Tuning these massive encoders could lead to overfitting.
  \item \textbf{Lack of diverse studies:} A recent review of 70 dermatological AI studies between 2015 and 2020 found that only 17 studies included ethnicity descriptors, and only 7 included skin tone descriptors \cite{daneshjou_2021_lack}. This could lead to under-specification of model performance for different ethnicities.
\end{itemize}

Denoising Diffusion Probabilistic Models (DDPMs) have been introduced \cite{ho_2020_denoising} as a new form of generative modeling. DDPMs have achieved state-of-the-art performance in image synthesis \cite{dhariwal_2021_diffusion} and are effectively applied in colorization \cite{song_2021_scorebased}, super-resolution \cite{saharia_2022_image}, segmentation \cite{baranchuk_2022_labelefficient}, and other tasks. In the medical domain, recent work has presented results for DDPM-based anomaly detection \cite{wolleb_2022_diffusion} and segmentation \cite{wu_2023_medsegdiffv2}, but these are limited to MRI, CT, and ultrasonography not natural smartphone-captured images of dermatology conditions. To our knowledge, none have explored segmentation and malignancy classification in this context from DDPM-based embeddings without re-training and evaluated performance on diverse dermatology images.

We introduce the FEDD framework, a denoising diffusion-based approach trained on small, skin tone-balanced, Diverse Dermatology Images (DDI) \cite{daneshjou_2022_disparities} subsets for skin lesion segmentation and malignancy classification that outperforms state-of-the-art across a diverse spectrum of skin tones and malignancy conditions with very few training examples. FEDD leverages the highly semantically meaningful feature embeddings learned by DDPMs for image synthesis. Finally, linear probes predict per-pixel class or per-image malignancy, achieving state-of-the-art performance on the DDI dataset without fine-tuning the encoder.

\section{Description of Data}

\begin{figure*}[t!]
    \centering
    \includegraphics[width=0.9\textwidth]{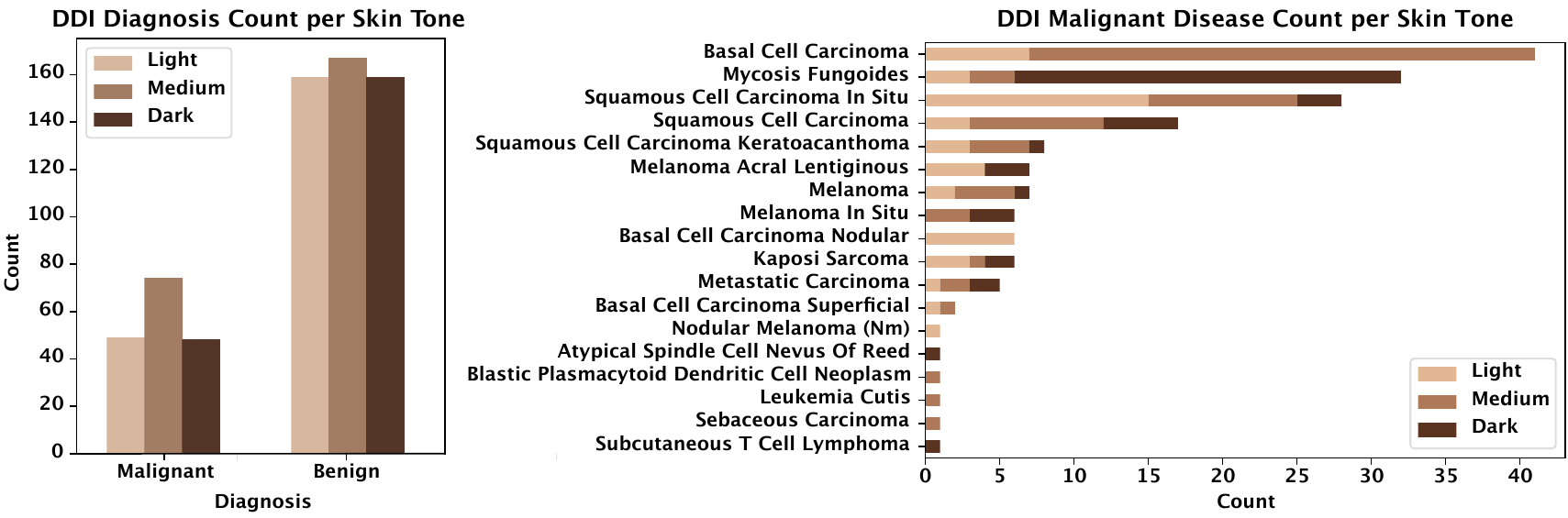}
    \caption{The disease count per skin tone (left) shows a smaller amount of malignancy data in DDI but otherwise mostly balanced between light and dark skin tones. The distribution of malignant illnesses (right) shows high diversity and thus the morphological variation of lesions in DDI.}
    \label{fig:DDI}
    \vspace{-2mm}
\end{figure*}

The most commonly used dataset to train and evaluate fairness in dermatology AI is Fitzpatrick17k \cite{groh_2021_evaluating,abid_2022_meaningfully,du_2023_fairdisco} thanks to its large size of nearly 17,000 images. However it contains a significant skin tone imbalance (3.6 times more light than dark skin toned samples) and greater than 30\% disease label noise \cite{groh_2021_evaluating}. Further, the samples are not biopsied and visual inspection itself can be an unreliable way of diagnosis without the use of histopathological information \cite{daneshjou_2022_checklist}. Recently, DDI \cite{daneshjou_2022_disparities} was published as a dermatological image dataset with Fitzpatrick-scale \cite{fitzpatrick_1988_the} scores for all images, classifying them as light (I-II), medium (III-IV), or dark (V-VI) in skin tone. While at a lower skin tone resolution (3-point instead of 6-point), these labels are reviewed by two board-certified dermatologists. It also includes a mix of rare and common benign and malignant skin conditions, all of which are confirmed via biopsy. The dataset contains visually ambiguous lesions that would be difficult to visually diagnose but represent the kind of lesions that are seen in clinical practice. DDI is somewhat balanced between skin tones, with about 16\% more information for medium skin tones; however, it is not balanced between malignant and benign classes. In total, the dataset contains 656 samples. Details and distribution of diagnosis are shown in Fig. \ref{fig:DDI}.

\begin{figure*}[t!]
    \centering
    \includegraphics[width=0.7\textwidth]{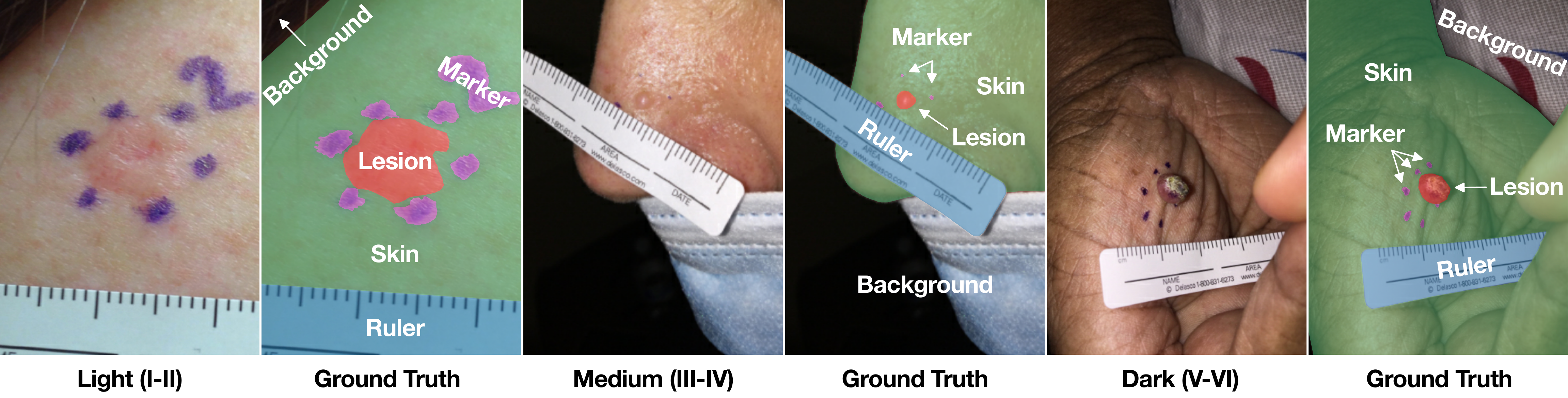}
    \caption{Ground truth segmentation samples. We color code lesions as red, skin as green, markings as purple, and rulers as blue. Backgrounds are left to be transparent.}
    \label{fig:DDI_annots}
    \vspace{-5mm}
\end{figure*}


We draw 4 balanced subsets of DDI for training, each representing approximately 5\% (10 samples per skin tone), 10\% (20 samples per skin tone), 15\% (30 samples per skin tone), and 20\% (40 samples per skin tone) of DDI. The smaller training sets are subsets of the larger ones, this is to say $5\% \subseteq 10\% \subseteq 15\% \subseteq 20\% \subseteq$ DDI. For classification we draw validation and test sets, each containing 30 samples (10 samples per skin tone). Further, we test model checkpoints trained on each DDI subset on all remaining DDI images (476 samples), accuracy results from this larger test set are reported on the paper text and on Table 3 of the supplementary materials. For segmentation, we test on 198 additionally annotated samples.


This is due to DDI including disease labels suitable for malignancy classification. For segmentation however, samples need to be semantically labeled and some samples may be difficult to correctly annotate, leading to discarding; for example if the target lesion is ambiguous, blurry, partially visible or occluded. We annotated the dataset and all masks underwent a secondary quality review. We define 5 classes: lesion, skin, marker, ruler, and background. We opted to label these classes as many images include a ruler or markings to denote the lesion of focus. Visualizations of our ground truth labels are shown in Fig. \ref{fig:DDI_annots}. Details on the annotation protocol, including skip criteria, can be found in the supplementary materials. We release our annotation work (a total of 378 annotated DDI images) as part of our contributions.

\section{Approach}
The UNet architecture was introduced for diffusion \cite{ho_2020_denoising} and found to improve generative performance \cite{jolicoeurmartineau_2020_adversarial} over other denoising score-matching architectures. Recent work \cite{dhariwal_2021_diffusion} has extensively ablated the diffusion UNet architecture by increasing depth while reducing the width, increasing the number of attention heads and applying it at different resolutions, applying BigGAN \cite{brock_2018_large} residual blocks, and introducing AdaGN for injecting timestep and class embedding onto residual blocks, obtaining state-of-the-art for image synthesis. From this work, we designate the unconditional image generation model as our backbone and freeze its weights. This network is an ImageNet-trained DDPM with $256\times256$ input and output resolution.
\begin{figure*}[t!]
    \centering
    \includegraphics[width=0.85\textwidth]{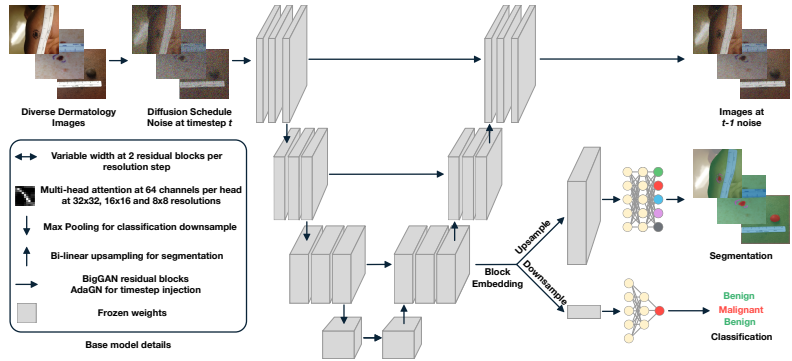}
    \caption{Image noise is added according to the diffusion noise schedule for the selected timestep. The DDPM processes the image, and feature embeddings are obtained from the desired block levels. Embeddings are concatenated and either up-sampled for segmentation or down-sampled for classification. Finally, Multi-Layer Perceptrons (MLPs) predict per-pixel semantic class or whole image malignancy.}
    \label{fig:FEDD}
    \vspace{-5mm}
\end{figure*}
\subsection{Lesion, Marker, Ruler, Skin, and Background Segmentation}
We obtain image encodings from blocks on the decoder side of the DDPM-UNet architecture, then apply bi-linear up-sampling up to some target output resolution ($256\times256$ in this case) and concatenate them before feeding them to MLPs for per-pixel classification following \cite{baranchuk_2022_labelefficient}. However, we use fewer blocks, a single-time step, and 5 MLPs. This is to avoid overfitting and because results from \cite{baranchuk_2022_labelefficient} describe how different blocks at different timesteps perform differently depending on the target data. To understand which blocks are most promising for dermatology images, we obtain a sample of feature encodings at different blocks and perform K-Means clustering shown in the supplementary materials. We selected the blocks which clustered semantically meaningful areas (e.g. lesion, skin, and ruler). We identify block 6 and block 8 at timestep 100 as the most promising and use this setup for the rest of our segmentation experiments.

\subsection{Malignancy Classification}

For classification, we down-sample block encodings using a combination of 2D and 1D max pooling operations until the feature vectors are one-dimensional and of size 512. We note that the total number of pooling operations varies depending on the sampled block, as deeper blocks are smaller with more channels, while shallower blocks are larger with fewer channels. The vectors are then passed onto a 3-layer MLP of size 64, 32, and 1. We include batch normalization and dropout between each layer with 50\% and 25\%. This classification network is trained to predict the malignancy of the input image from the down-sampled feature vector. A summary of our approach is shown in Fig. \ref{fig:FEDD}.

\section{Results and Discussion}
\label{sec:results}

\subsection{Lesion, Marker, Ruler, Skin, and Background Segmentation}

Our segmentation results are evaluated by Intersection over Union (IoU) performance and are compared against other architectures pre-trained on ImageNet: DenseNet121 \cite{huang_2016_densely}, VGG16 \cite{simonyan_2014_very}, ResNet50 \cite{he_2015_deep}, and two other smaller networks, EfficientNetB0 \cite{tan_2019_efficientnet} and MobileNetV2 \cite{sandler_2018_mobilenetv2}. These architectures are configured as UNets and tasked with segmenting the input images. We observe FEDD outperforms all other architectures across all subsets of DDI on our validation and test sets. Importantly, other architectures (particularly the smaller networks) close the performance gap as the amount of training data increases, showcasing that FEDD's efficiency is most prevalent in very small data scenarios.

\begin{figure*}[t!]
    \centering
    \includegraphics[width=1\textwidth]{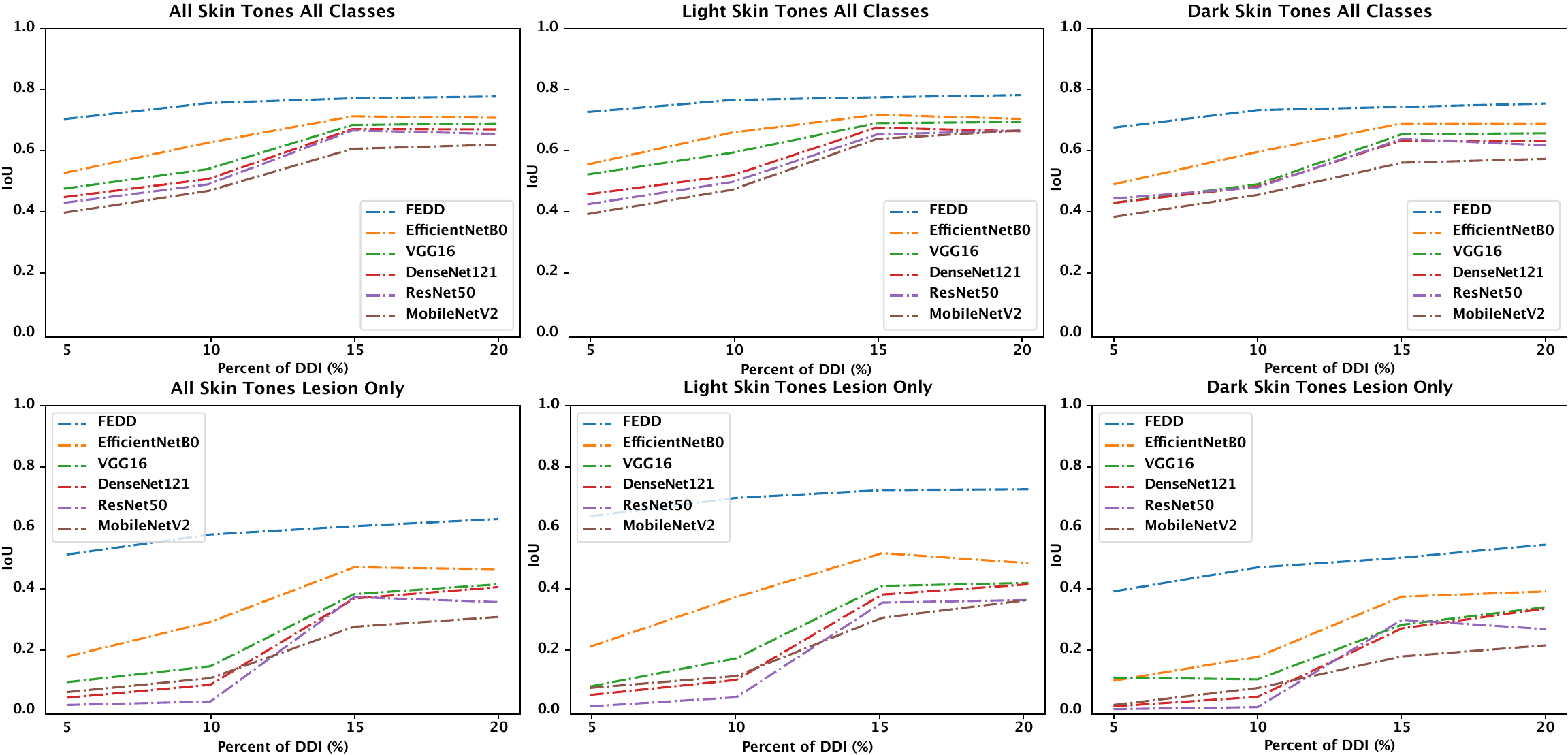}
    \caption{\textbf{Top Row}: The test IoU score for all segmentation classes for all, light, and dark skin tones (left-to-right). \textbf{Bottom Row}: The test IoU score for the lesion segment for all, light, and dark skin tones (left-to-right).}
    \label{fig:IoUs}
    \vspace{-5mm}
\end{figure*}

We further compare the FEDD's IoU performance on light and dark skin tone images to showcase algorithmic fairness. We note that all architectures show similar performance for both skin tones, suggesting that our balanced data subsets play a larger role in fairness than the choice of neural network. Finally, we plot test set performance when only considering the lesion class split between light and dark tones. We find that FEDD's performance is significantly better at segmenting the lesion class compared to other architectures. We believe this is due to the greater morphology variation of different skin lesions being harder to learn than other more consistent targets like the ruler. Since DDI contains a diversity of skin conditions, FEDD's efficiency becomes very useful for high-quality lesion segmentation of this morphologically changing class. These results are shown in Fig. \ref{fig:IoUs}.

\begin{figure*}[t!]
    \centering
    \includegraphics[width=0.85\textwidth]{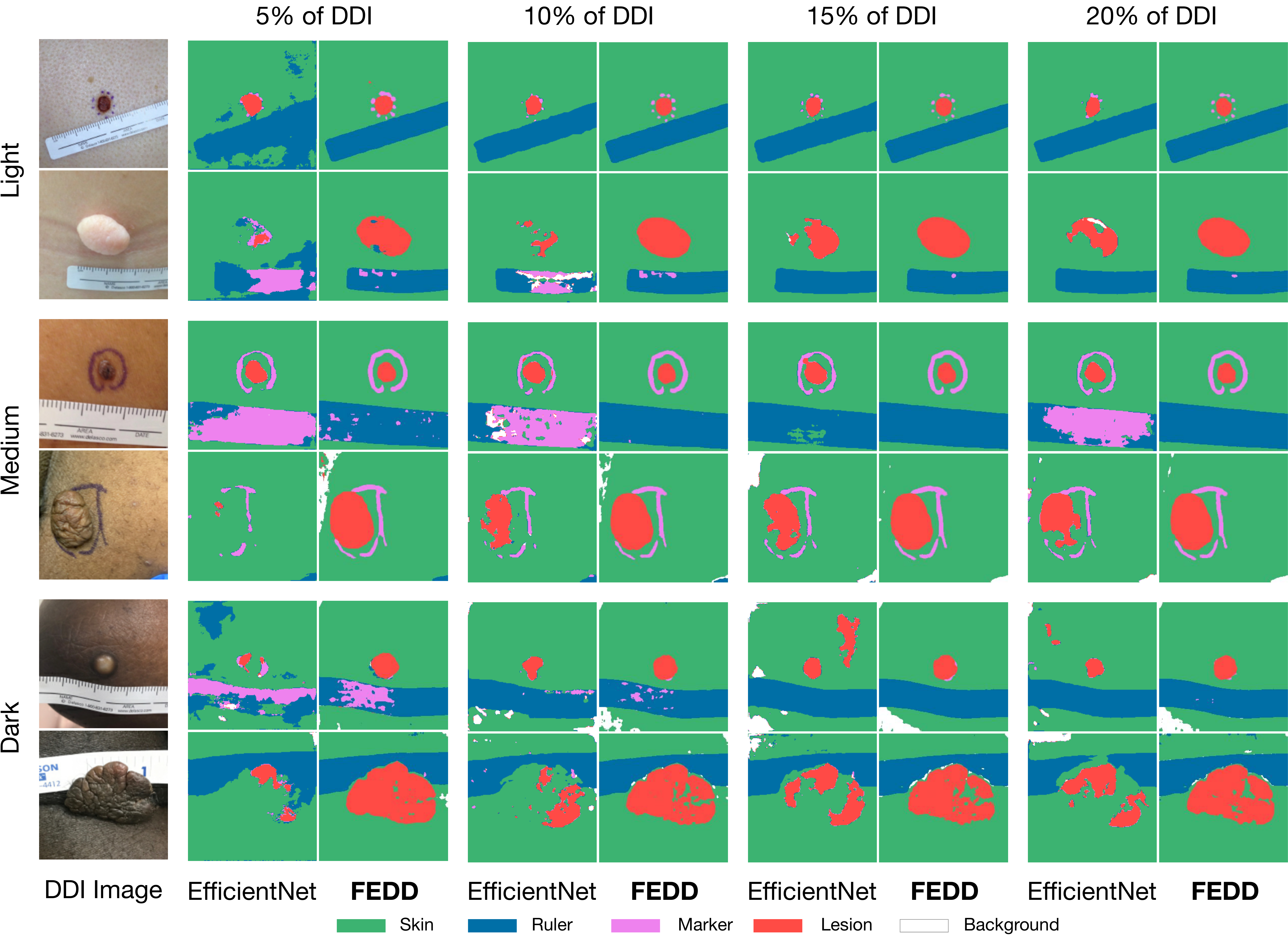}
    \caption{Test set segmentation results for FEDD and EfficientNet.}
    \label{fig:Segs}
    \vspace{-5mm}
\end{figure*}

In Fig. \ref{fig:Segs}, we visualize predicted segmentation masks between FEDD and the next best IoU-performing architecture, EfficientNetB0. We observe that FEDD achieves significantly better segmentation masks at lower fractions of labeled data across all skin tones. With a larger percentage of labeled data, EfficientNet begins to produce similar results to FEDD, but FEDD comparatively outputs higher-quality segmentations with fewer segmentation artifacts and false positives. The skin lesions themselves, which appear in different sizes, locations, and morphologies, are also most accurately segmented by FEDD.

\subsection{Malignancy Classification}

\begin{figure*}[t!]
    \centering
    \includegraphics[width=1\textwidth]{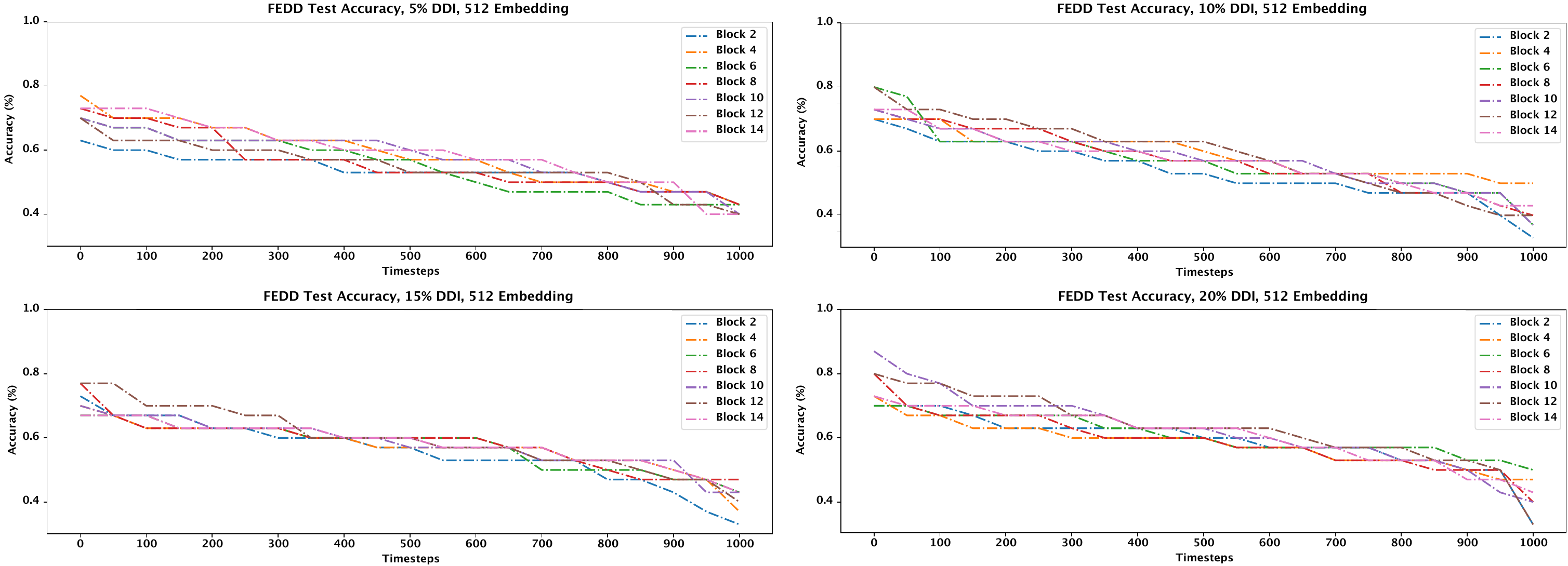}
    \caption{Classification accuracy of each DDPM UNet decoder is shown. Later steps in the reverse diffusion process produce the highest quality embeddings. When less data is available (top row), earlier blocks of the UNet decoder perform best. When more data is available (bottom row), the later, shallower blocks of the decoder perform best.}
    \label{fig:Cls_Blocks}
    \vspace{-2mm}
\end{figure*}

We ablate the performance of each individual block per timestep in the classification context as, to the best of our knowledge, it has not been done before. It is also not entirely intuitive which block depth and timestep combination would produce the best representations for classification, as well as how that performance varies as we introduce more data. We train FEDD's classifier on block embeddings produced between timestep 0 and 1000 of the backward diffusion process. We then record the accuracy of the classifier per block in increments of 50 timesteps. Fig. \ref{fig:Cls_Blocks} describes these results. While noisy, given the small amount of test data, the general pattern is that the earlier time-steps (later in the reverse diffusion de-noising process) allow for higher classification accuracy. This is likely due to the quality of the DDPM sample increasing as more noise is removed. Another finding is that as the classifier is shown more data, the shallower blocks begin to perform better. The best performing blocks are shown to be: block 4 at 5\% DDI, block 6 at 10\% DDI, block 12 at 15\% DDI, and block 14 at 20\% DDI. We attribute this to the fact that shallower blocks of the UNet decoder capture finer detail of the reconstructed image while deeper blocks capture lower-resolution detail. This coarse data is more generic and thus more generalizable than the finer features in later blocks. As we increase the amount of data, the classifier has enough information to learn from the finer details of later blocks, boosting performance.

\begin{figure*}[t!]
    \centering
    \includegraphics[width=1\textwidth]{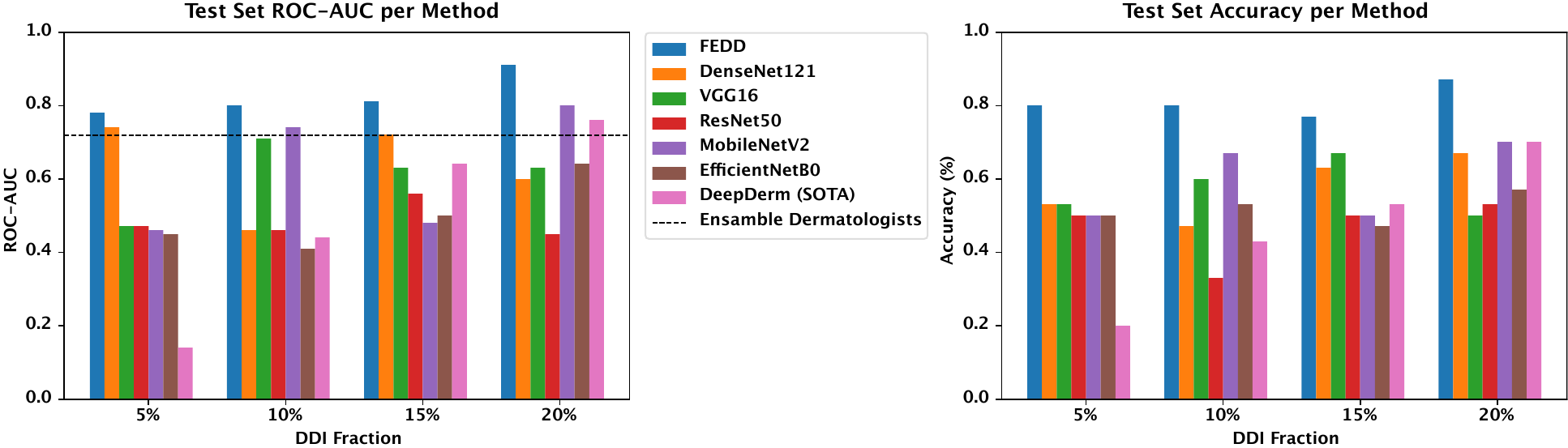}
    \caption{ROC-AUC scores (left) show FEDD outperforms other methods, including an ensemble of dermatologists. Accuracy per method is also higher (right).}
    \label{fig:roc_auc_acc}
    \vspace{-5mm}
\end{figure*}

We select the best-performing block and timestep combination at each fraction of data for the rest of our experiments. The previous classification state-of-the-art on the DDI dataset is reported on \cite{daneshjou_2022_disparities} as the DeepDerm \cite{esteva_2017_dermatologistlevel} architecture pre-trained on HAM10000 \cite{tschandl_2018_the} and fine-tuned on DDI. We compare FEDD performance against this setup as well as other commonly used classifiers on each of our DDI subsets. We measure Receiver Operating Characteristic Area Under the Curve (ROC-AUC) at the best threshold for each method, F1 scores, and classification accuracy. We observe FEDD obtains a higher ROC-AUC than any other method at every level of data. It also surpasses the dermatologist ensemble performance reported by \cite{daneshjou_2022_disparities}. FEDD also reports the best accuracy, however, it does not see improvement at 10\% or 15\% of DDI compared to 5\% and 20\%, as shown in Fig \ref{fig:roc_auc_acc}. When observing ROC curves for FEDD, we see it meets or exceeds the ensemble of dermatologists even at the smallest subset of DDI. We divide F1 scores between the light and dark skin tones finding that FEDD does not always obtain the best F1 performance at larger subsets of data, namely 15\% and 20\% of DDI. This result suggests that purpose-built classification networks could have a performance advantage over diffusion embeddings applied toward classification when allowed to train on larger amounts of data. Detailed F1 results are shown in table form on the supplementary materials.


\section{Conclusion}
We introduce the FEDD framework for skin lesion segmentation and malignancy classification that outperforms state-of-the-art methods and an ensemble of board-certified dermatologists across a diverse spectrum of skin tones and malignancy conditions under limited data scenarios. Our proposed methodology can improve the diagnosis and treatment of skin diseases while maintaining fair segmentation outcomes for under-represented skin tones and accurate malignancy predictions for rare malignancy conditions. We freely release our code and annotations to encourage further research around dermatological AI fairness.
\bibliographystyle{IEEEtran}
\bibliography{ref}

\end{document}


\title{Supplementary Materials for FEDD - Fair, Efficient, and Diverse Diffusion-based Lesion Segmentation and Malignancy Classification}
\titlerunning{FEDD}
%
\author{Héctor Carrión$^{1*}$, Narges Norouzi$^{2}$}

%
\authorrunning{Carrión et al.}
%
%
\institute{University of California, Santa Cruz$^1$, University of California, Berkeley$^2$,
\email{hcarrion@ucsc.edu$^*$}} 
%
%
%
\maketitle              
%

%
%
%

\section{Dataset and annotation details}

As mentioned on the main text, we draw 4 balanced subsets of DDI for training, each representing approximately 5\% (10 samples per skin tone), 10\% (20 samples per skin tone), 15\% (30 samples per skin tone), and 20\% (40 samples per skin tone) of DDI. The smaller training sets are subsets of the larger ones, this is to say $5\% \subseteq 10\% \subseteq 15\% \subseteq 20\% \subseteq$ DDI. For classification we draw validation and test sets, each containing 30 samples (10 samples per skin tone). Further, we test model checkpoints trained on each DDI subset on all remaining DDI images (476 samples), accuracy results from this larger test set are reported on the paper text and on supplementary materials Table 3. For segmentation we test on 198 additionally annotated samples (59 light, 80 medium and 59 dark in skin-tone). These larger test sets are skin-tone unbalanced, as expected in clinical settings. All validation and testing sets are disjoint from the training splits and from each other.

For malignancy classification, DDI includes disease labels. For segmentation, sample images need to be semantically labeled and some may be difficult to annotate; the annotation protocol is as follows:

\begin{enumerate}
  \item Lesion is segmented following the boundary at which the skin transitions from healthy to unhealthy appearance.
  \item Markings or rulers are segmented in their visible totality.
  \item Non-lesion, or normal skin is segmented.
\end{enumerate}

This denoted our segmentation masks which cover 5 different classes: lesion, skin, marker, ruler, and background. We opted to label these classes as many images include a ruler and markings to denote the lesion of focus. All other potentially present objects, like clothing, were denoted as background.

Not all images were suitable for labeling as DDI images may not have been collected with computer vision in mind and consequently were skipped. The following criteria will trigger a skip: 

\begin{enumerate}
  \item Lesion is occluded.
  \item Lesion is significantly blurry.
  \item Lesion is only partially visible.
  \item Target lesion is ambiguous (no lesions marked when multiple are present or multiple are marked in a single image).
  \item Lesion is on or near the scalp (hair is not a labeled target and time consuming to annotate).
\end{enumerate}

Examples of images which triggered a skip can be found on DDI sample id 25, 55, and 161. The total number of annotated images which passed our annotation criteria and secondary review is 378. We have released this annotation work on our github linked in the abstract. Table \ref{tab:DDI_Subsets} describes our training subsets of the Diverse Dermatology Images dataset.

\begin{table}[h]
\centering
\caption{The distribution of training data per sub-set of DDI.}
\resizebox{1\textwidth}{!}{
\begin{tabular}{||c|c|c|c|c||}
    \hline
    DDI Subset & Total Samples & Samples per Skin Tone & Malignant Samples per Skin Tone & Benign Samples per Skin Tone\\
    \hline
    5\% & 30 & 10 & 5 & 5 \\
    \hline
    10\% & 60 & 20 & 10 & 10 \\
    \hline
    15\% & 90 & 30 & 15 & 15 \\
    \hline
    20\% & 120 & 40 & 20 & 20 \\
    \hline
\end{tabular}
}
\label{tab:DDI_Subsets}
\end{table}

\section{Software and Hardware}

\subsection{Setup}

All code involving this study was written in Python v3.7.12. The packages used for training and inference were the latest available stable versions of deep learning frameworks PyTorch v1.13, Keras v2.8, and TensorFlow v2.8. Additional packages used for numerical processing, plotting and visualizations were Pandas v1.3.5, NumPy v1.21.5, sklearn v1.0.2, PIL 7.1.2, and Matplotlib 3.5.1. The hardware used was a Google Colab instance running a quad-core Intel Xeon CPU at 2.3 GHz with 80GB of RAM and an NVIDIA A100 GPU with 40GB of vRAM.

\subsection{Performance}

The average run-time for FEDD training was about 40 minutes per MLP over 100 epochs. Memory footprint did not exceed 80GB of system memory or 40GB GPU memory after multiple runs. Inference time is equal to about 1.5 images per second.

\section{Additional Hyper-parameters and Experimental Setup}

Additional hyper-parameters used were 30 batch size, 60 batch size, 90 batch size, and 120 batch size for 5, 10, 15, and 20\% of DDI, respectively. These were selected as they allowed all subset data to be processed in a single batch. For testing and validation, a 30-batch size was used. The Adam optimizer was used, along with weight decay equal to $0.00001$. The loss function is Binary Cross Entropy. The MLPs were trained over 500 epochs with early stopping monitoring validation loss. Most runs early-stopped at or before 100 epochs.

\section{K-Means Clustering}

\begin{figure*}[h!]
    \centering
    \includegraphics[width=1\textwidth]{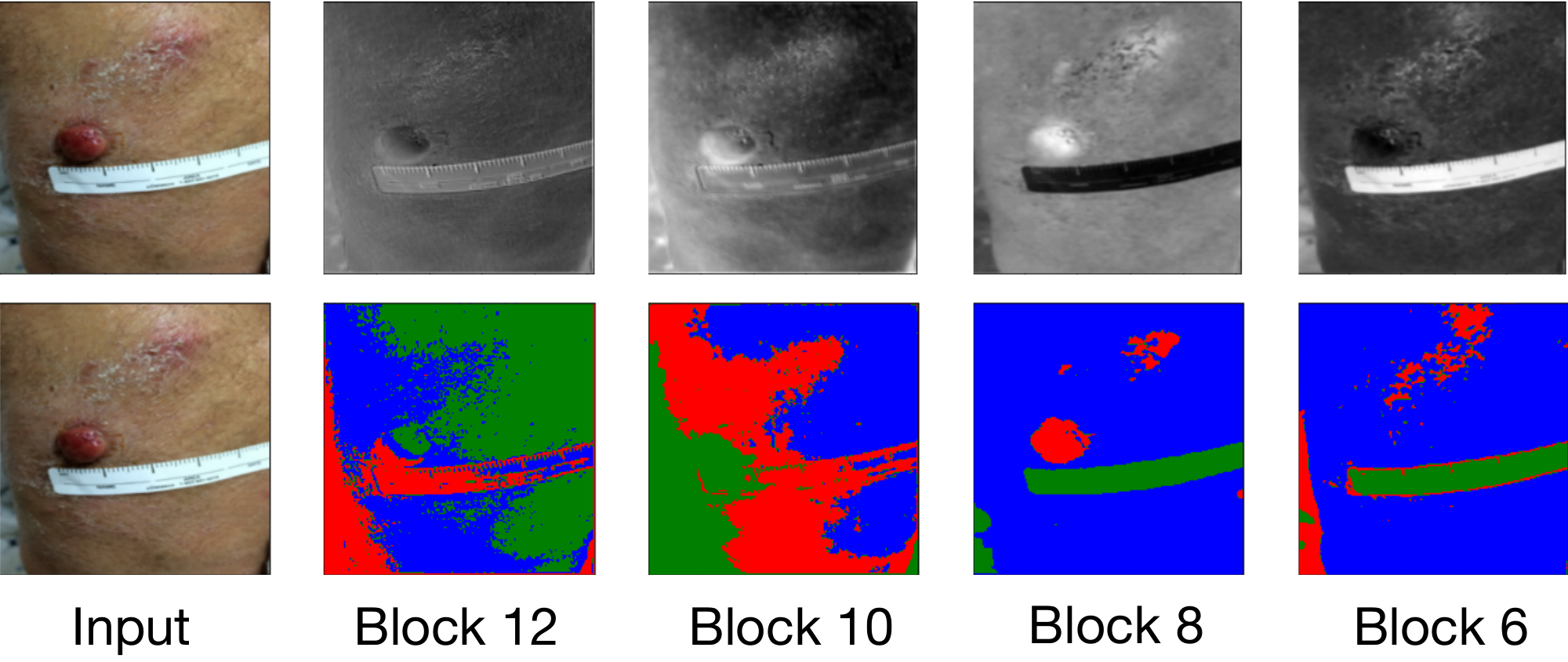}
    \caption{FEDD activation maps (top) and their clustering quality (bottom).}
    \label{fig:KMeans}
\end{figure*}

Figure \ref{fig:KMeans} represents some examples of clustering quality for FEDD block activations and embeddings. K-Means was used with $K=3$.

\section{ROC Curves}

\begin{figure*}[h!]
    \centering
    \includegraphics[width=1\textwidth]{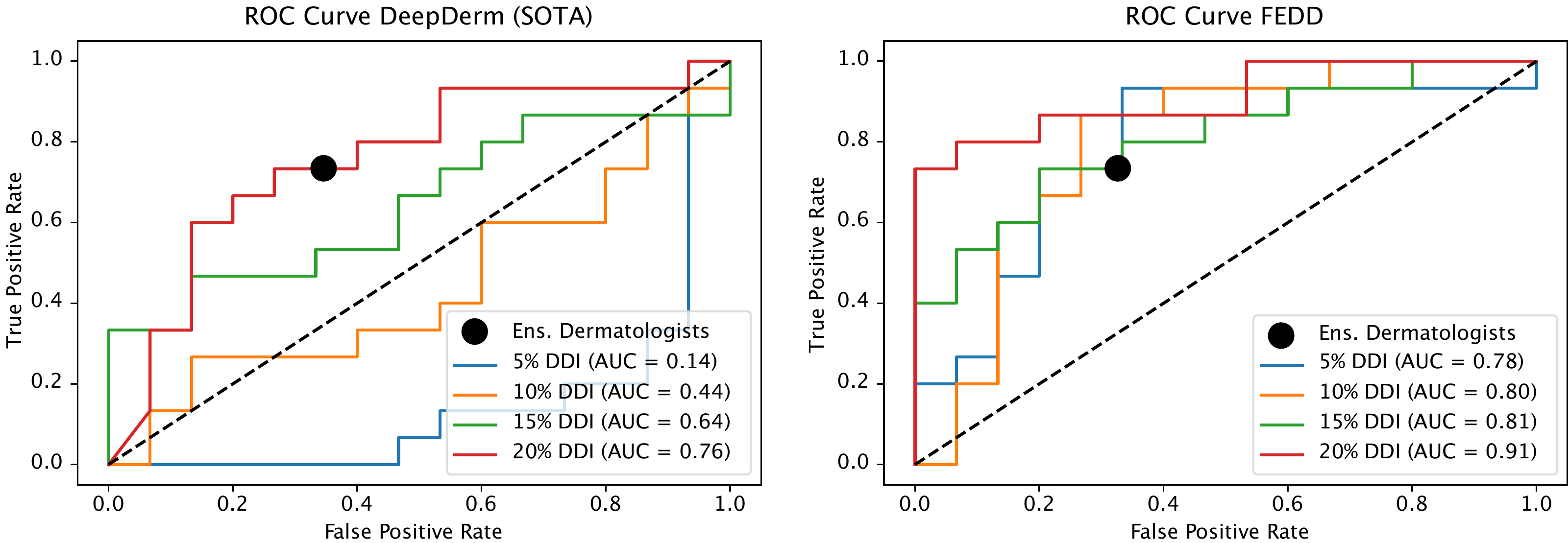}
    \caption{Test set ROC Curve for DeepDerm (left) and FEDD framework (right) at each subset of DDI data. The performance of an ensemble of dermatologists is also shown as a black circle.\\}
    \label{fig:roc}
    \vspace{-5mm}
\end{figure*}

Figure \ref{fig:roc} showcases the ROC curves for previous state-of-the-art and FEDD.

\section{F1 Performance per Skin Tone}

Table \ref{tab:F1} demonstrates the F1 performance per skin tone for FEDD and all other tested architectures.

\section{Improvement Deltas}

Table \ref{tab:Deltas} demonstrates the performance advantage of FEDD versus the next best tested method.

\\
\begin{table}[h!]
\centering
\caption{F1 Performance for all tested architectures split between light (left) and dark (right) skin tones.}
\resizebox{1\textwidth}{!}{
\begin{tabular}{||c|c|c|c|c|c|c|c|c||}
    \hline
    Method & 5\% DDI (Light) & 10\% DDI (Light) & 15\% DDI (Light) & 20\% DDI (Light) & 5\% DDI (Dark) & 10\% DDI (Dark) & 15\% DDI (Dark) & 20\% DDI (Dark)\\
    \hline
    DenseNet121 & 0.60 & 0.62 & 0.57 & 0.67 & \textbf{0.80} & 0.71 & 0.57 & 0.77 \\
    \hline
    VGG16 & 0.62 & 0.50 & \textbf{0.73} & 0.73 & 0.67 & 0.57 & 0.62 & 0.77 \\
    \hline
    ResNet50 & 0.00 & 0.25 & 0.29 & 0.00 & 0.33 & 0.60 & 0.50 & 0.00 \\
    \hline
    EfficientNetB0 & 0.67 & 0.00 & 0.29 & 0.62 & 0.46 & 0.00 & 0.44 & \textbf{0.91} \\
    \hline
    MobileNetV2 & 0.55 & 0.33 & 0.62 & \textbf{0.89} & 0.67 & 0.75 & 0.55 & 0.89 \\
    \hline
    DeepDerm & 0.67 & 0.00 & 0.44 & 0.60 & 0.67 & 0.50 & 0.57 & 0.75 \\
    \hline
    \textbf{FEDD} & \textbf{0.73} & \textbf{0.83} & 0.67 & \textbf{0.89} & \textbf{0.80} & \textbf{0.83} & \textbf{0.75} & 0.89 \\
    \hline
\end{tabular}
}
\label{tab:F1}
\end{table}

\begin{table}[h!]
\centering
\caption{Performance Improvement of FEDD compared to next-best method on the full segmentation and classification test sets (skin-tone unbalanced).}
\begin{tabular}{||c|c|c|c|c||}
    \hline
    Metric & 5\% DDI & 10\% DDI & 15\% DDI & 20\% DDI \\
    \hline
    IoU & 0.18 & 0.13 & 0.06 & 0.07 \\
    \hline
    Accuracy & 14\% & 6\% & 5\% & 4\% \\
    \hline
\end{tabular}

\label{tab:Deltas}
\end{table}
\section{Expanded results visualization}

\begin{figure*}[t!]
    \centering
    \includegraphics[width=1\textwidth]{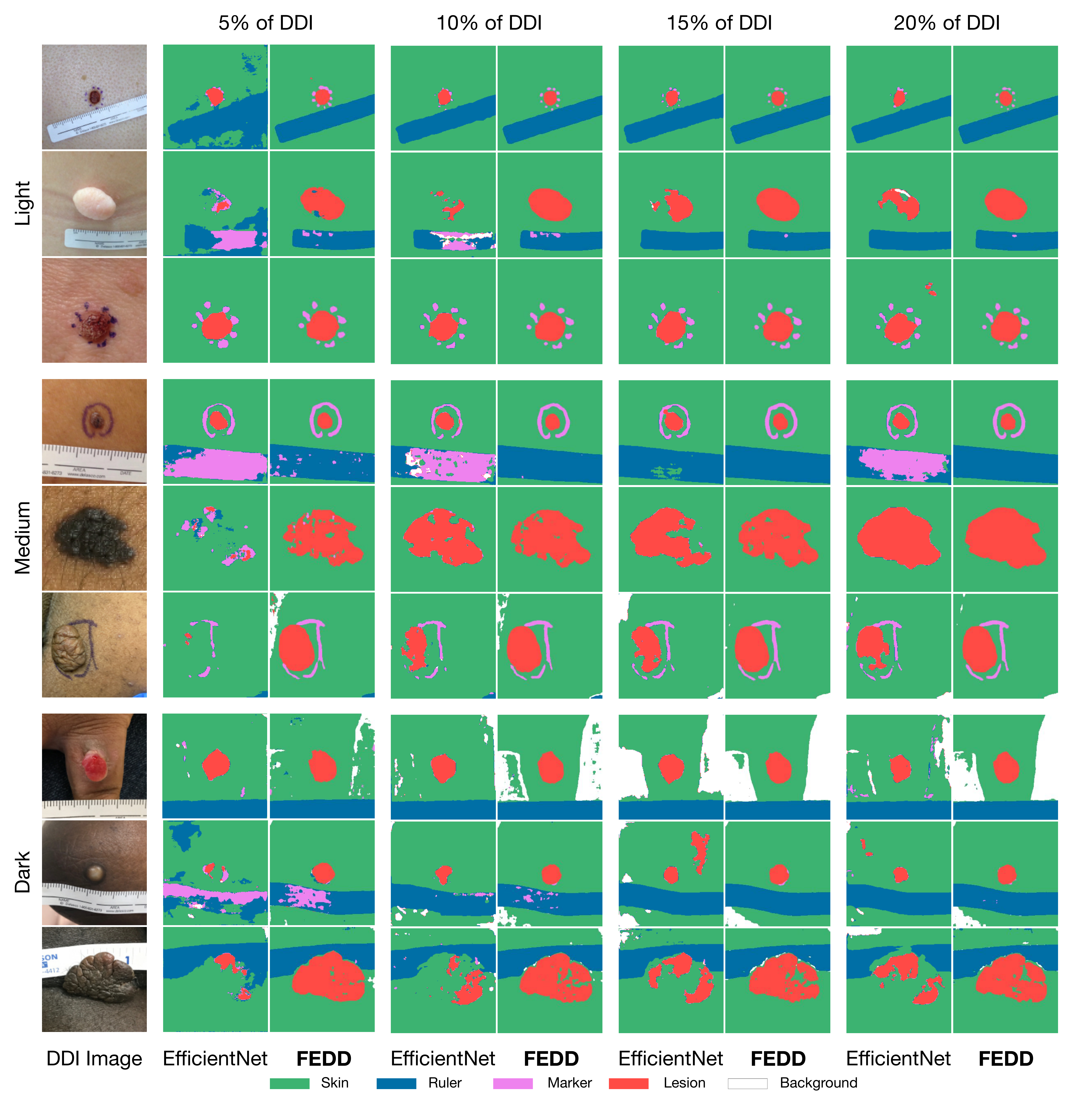}
    \caption{Expanded test set segmentation results for FEDD and EfficientNet.}
    \label{fig:Segs_2}
    \vspace{-2mm}
\end{figure*}

Fig \ref{fig:Segs_2} showcases additional segmentation results in full-width for easy viewing.

